\documentclass{article}

\usepackage[final]{neurips_2020}

\usepackage[utf8]{inputenc} 
\usepackage[T1]{fontenc}    
\usepackage{hyperref}       
\usepackage{url}            
\usepackage{booktabs}       
\usepackage{amsfonts}       
\usepackage{amsmath, amsthm, amssymb}

\usepackage{nicefrac}       
\usepackage{microtype}      
\usepackage{graphicx}
\usepackage{enumitem}

\title{Unsupervised Data Augmentation for Object Detection}

\author{%
  Zhang Yichen \\
  School of Computing \\
  National University of Singapore \\
  \texttt{zhang.yichen@u.nus.edu} \\
  \And
  Song Zeyang \\
  School of Computing \\
  National University of Singapore \\
  \texttt{zeyang\_song@u.nus.edu} \\
  \AND
  Li Wenbo \\
  School of Computing \\
  National University of Singapore \\
  \texttt{wenbo@u.nus.edu} \\
}

\begin{document}

\maketitle

\begin{abstract}
  Data augmentation has always been an effective way to overcome overfitting issue when the dataset is small. There are already lots of augmentation operations such as horizontal flip, random crop or even Mixup. However, unlike image classification task, we cannot simply perform these operations for object detection task because of the lack of labeled bounding boxes information for corresponding generated images. To address this challenge, we propose a framework making use of Generative Adversarial Networks(GAN) to perform unsupervised data augmentation. To be specific, based on the recently supreme performance of YOLOv4, we propose a two-step pipeline that enables us to generate an image where the object lies in certain position. In this way, we can accomplish the goal that generating an image with bounding box label. 
\end{abstract}

\section{Introduction}

Computer Vision is one of the most important areas in Artificial Intelligence. It contains several basic tasks: image classification \cite{2015arXiv151203385H}, object detection \cite{2013arXiv1311.2524G}, instance segmentation \cite{2017arXiv170306870H} and semantic segmentation \cite{2014arXiv1412.7062C}. Among them, object detection has attracted widespread attention from researchers all over the world. It solves two main problems in real world applications: "What is the object?" and "Where is the object?". Current methods are simulating human vision and cognition in specific application scenarios, such as pedestrian detection, face detection and text detection. In recent years, with the development of deep learning technology, significant breakthroughs have been made in object detection, and it has also been applied to more fields such as automatic driving, robot vision, and video surveillance.

Although object detection has developed rapidly and new algorithms have constantly refreshing the scores of major rankings, one of its shortcomings is its difficulty in obtaining labeled data compared to classification task because of the tedious bounding box annotations process. A straight forward solution to this is generating pseudo label from trained detection model with confidence directly. However, it seems useless or even harmful since it's possible that the trained model will make mistake even if it gives out high confidence value.

During the survey, we found that some methods start using Generative Adversarial Network (GAN) for data augmentation recently, which achieve better results than traditional methods. For example, DAGAN \cite{antoniou2018data} succeed in generating augmented images for human face as shown in Figure \ref{DAGAN}, where only the left top one is real image. And after training on those augmented images, their model perform much better compared to those methods only using the real images. Inspired by these methods, we raise an enlightening question about this.

\begin{itemize}[leftmargin=*]
    \item What if we can design a framework about GAN to generate augmented image together with the bounding box labels? 
\end{itemize}

To answer the above question, we develop a framework aiming to perform data augmentation for object detection based on the success of YOLOv4 and GAN. To be specific, we propose a two-step framework. In first step, our goal is to train a GAN for normal data augmentation based on original dataset, i.e., be able to generate real images. In second step, our goal is further extended to generate images where object should lies in certain position. And we use YOLOv4 to localize the position of object in the generated images and thus instruct the training of generator. With this model, the only thing we need to do is providing a position to the generator, then we can get the generated images that meet the requirement. Further more, we can use this technique to generate some hard samples, such as images with small or bias object. With these hard samples, model may become more robust after training.

In summary, the contributions of this paper are three folds:
\begin{itemize}
    \item Propose a novel data augmentation framework to solve the essential problem of lacking labeled sample in object detection.
    \item Encode the position constraint into the generated images.
    \item Provide comprehensive analysis and discussion on our results and experiment process, which should be useful to other people who are also new to this area.

\end{itemize}

To our best knowledge, this is the first work that applying GAN to generate the whole images directly based on a pure condition for object detection task.

\section{Related Work}

    \subsection{Object Detection}
        
        Most of the early object detection methods were constructed based on manual features. Due to the lack of effective image representation at that time, people could only design complex feature representations. A well-known method is the Viola Jones detector \cite{990517}. The VJ detector uses a sliding window to view all possible window sizes and positions in an image. With the performance of manual feature selection technology becoming saturated, object detection has reached a stable development period after R. Girshick et al. proposed a region with CNN features (RCNN) for object detection in 2014 \cite{7112511}. Since then, object detection has developed at an unprecedented speed. In the era of deep learning, object detection can be divided into two categories: "two-stage detection" and "one-stage detection".
        
        The former defines the detection frame as a "from coarse to fine" process, the most typical methods are RCNN \cite{vandeSandeICCV2011}, Fast RCNN \cite{2015arXiv150408083G}, Faster RCNN \cite{NIPS2015_14bfa6bb}, and later also proposed a variety of improvements, including RFCN and Light head RCNN.
        
\begin{figure}[htbp]
        \centering
        \begin{minipage}[t]{0.48\textwidth}
            \centering
            \includegraphics[scale=0.35]{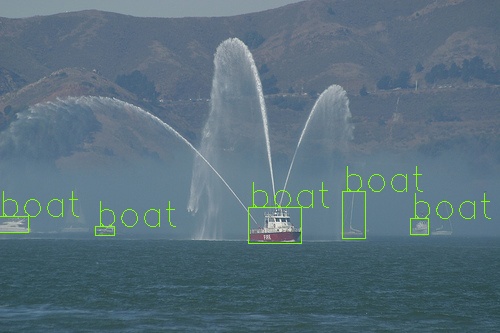}
            \caption{YOLOv4 Detection Result}
            \label{yolov4}
        \end{minipage}
        \begin{minipage}[t]{0.48\textwidth}
        \centering
            \includegraphics[scale=0.4]{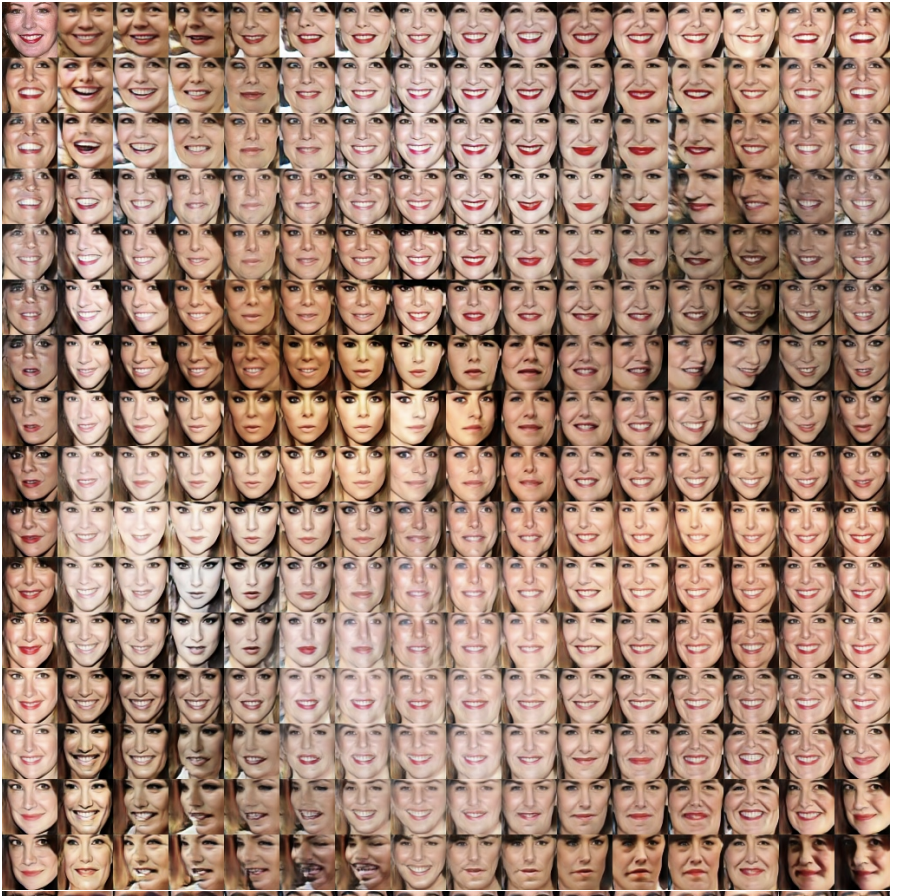}
            \caption{Outcome of DAGAN}
            \label{DAGAN}
        \end{minipage}
        \end{figure}

        The latter defines it as "one step in place", and typical methods include YOLO, SSD \cite{2015arXiv151202325L}, and RetinaNet \cite{2017arXiv170802002L}. We used the YOLOv4 \cite{2020arXiv200410934B} framework this time, the backbone used is the improved CSPDarknet \cite{2019arXiv191111929W} and the new feature map usage method BiFPN \cite{2019arXiv191109070T} is used, both of which are robust methods in the field of object detection.In Figure \ref{yolov4}, We can see that even difficult sample containing small objects, YOLOv4 can still provide excellent results. So we can rely it on generating pseudo bounding box labels to some degree.

    \subsection{Generative Adversarial Network}
        GAN was first proposed by Ian Goodfellow in 2014. By introducing the concept of adversarial learning, generator and discriminator trying their best to compete with each other and gradually both become better during this process. Because of its success in generating samples, GAN has been paid more and more attention by researchers in recent years and has been applied to different fields such as style transfer, portrait generation (StyleGAN \cite{2018arXiv181204948K}), few-shot, one-shot, and zero-shot learning (AFHN \cite{2020arXiv200313193L}). 
        
        GAN usually consists of a generator and a discriminator, where the generator needs to generate fake samples to fool the discriminator while the discriminator needs to have the ability to distinguish between real and fake samples.

         
         However, GAN has suffered from lots of problems such as model collapse, difficulty in finding a balance between generator and discriminator, and the lack of variety in generated samples. Many papers have tried to solve the problem, but the effect is not satisfactory and the problem has not been completely solved.
         
         To solve the balance problem between generator and discriminator, which may cause model collapse, Wasserstein GAN (WGAN) \cite{arjovsky2017wasserstein} introduce gradient penalty and Wasserstein distance, while the latter can also act as the indicator for the training progress. It also has some other simple modifications in GAN's structure such as removing logarithm from the loss function, removing the final Sigmoid layer of discriminator and outputs the score of input samples rather than probability. All these simple changes largely solve the above mentioned problems of GAN. 
        

    \subsection{Data Augmentation}
        
        In object detection, traditional data augmentation methods include random crop, changing picture size, changing picture tone, and random clip. Based on the idea of traditional methods, this year there are also many outstanding improvement methods such as mosic \cite{2020arXiv200410934B}, mixup \cite{2017arXiv171009412Z}, roimix \cite{2019arXiv191103029L}. The above-mentioned methods all transform the original image to the augmented data. They can be well explained, but most of them are not effective and cannot generate some unseen 
        images.
        
        
        Besides these traditional methods, people also use deep learning methods, such as GAN, for data augmentation. ASDN\cite{2017arXiv170403414W} use masks in the feature layer to block some parts and generate images without some important features. This method improves the network's utilization of various features. MTGAN\cite{Bai_2018_ECCV} use data augmentation for small objects detection, and they use GAN to expand the resolution for small objects with low pixels. This can improve the recognition accuracy of small objects, but both methods do not expand the dataset. AugGan\cite{huang2018auggan} converts images in daytime to other time like night, which could improve generalization ability of object detection model. However, this model requires semantic segmentation information of input samples, which is hard to acquire in ordinary datasets.

\section{Methodology}
    In this section, we will talk about our overall framework and its possible useful application.

    \subsection{Overall Framework}
    \begin{figure}
          \centering
          \includegraphics[scale=0.4]{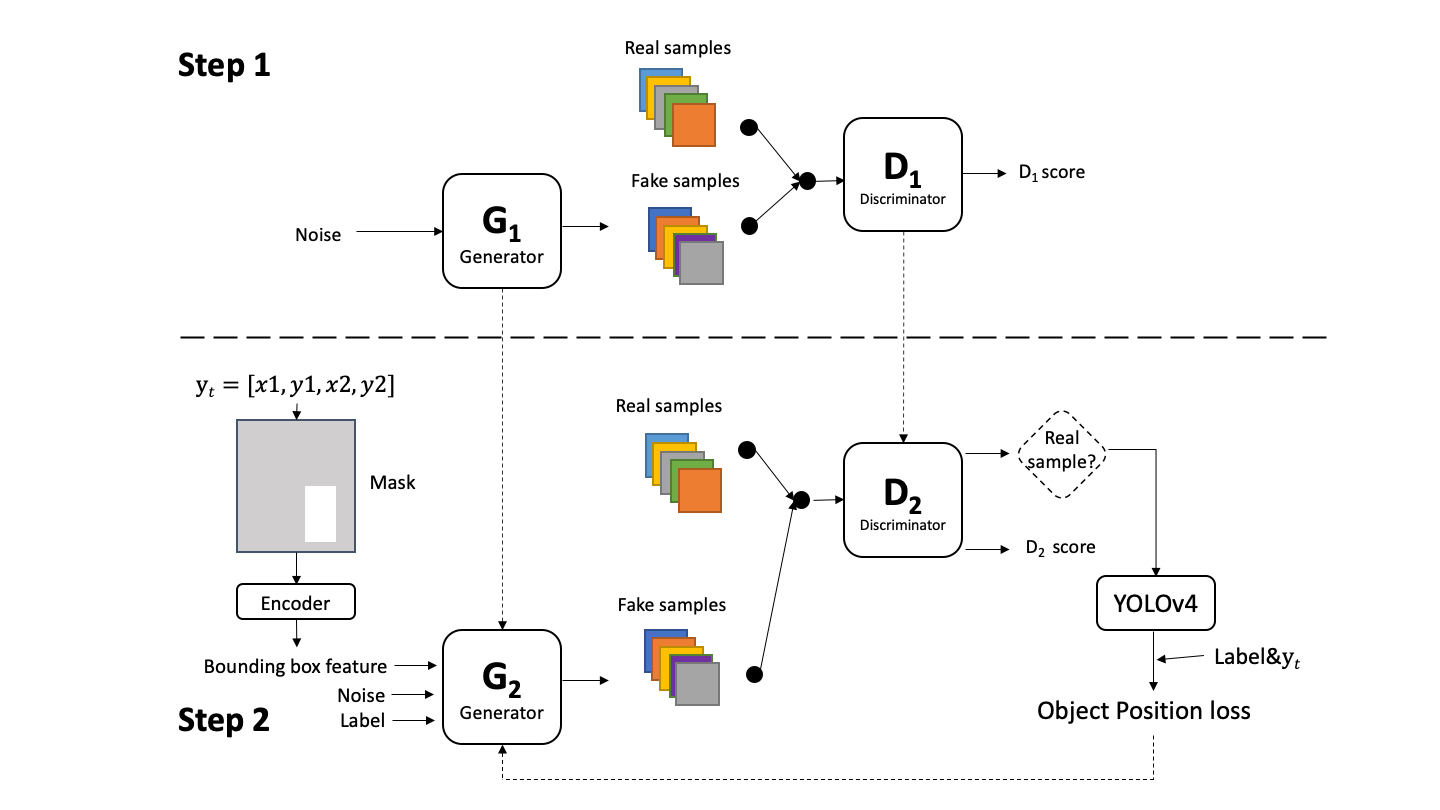}
           \caption{Framework}
           \label{framework}
    \end{figure}
    Our model is proposed to generate visually plausible images with objects in certain position.
    For this purpose, we design a 2-step training framework as shown in Figure \ref{framework}. 

    \paragraph{Step 1}
         In the first step, in order to generate indistinguishable fake images, 
         we first train a GAN which is able to translate noise to the target image domain. In this part, we adopt Wasserstein GAN structure for the framework. 
         
\begin{figure}
          \centering
          \includegraphics[scale=0.4]{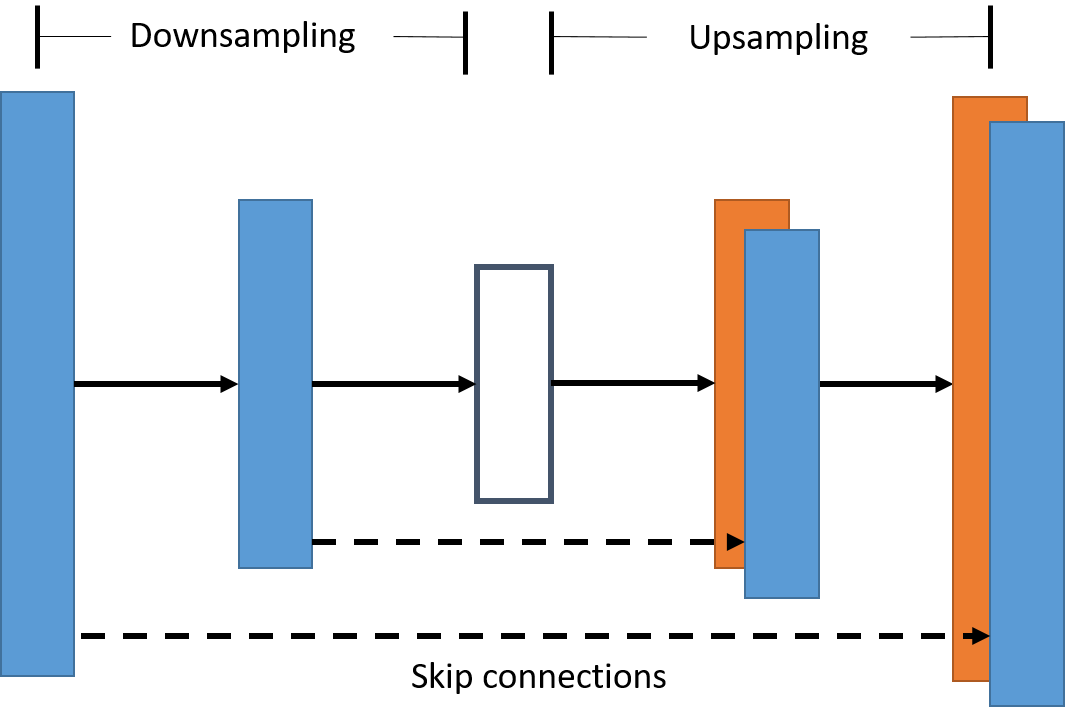}
          \caption{Unet structure used in our generator}
          \label{Unet} 
          \end{figure}

         In this framework, there are two main components: generator and discriminator. For generator, we used a network following the general architecture of U-Net\cite{ronneberger2015unet} as shown in Figure \ref{Unet}). Then for the discriminator, a straight forward convolutional neural network is applied to predict the score of input samples. Note that CNN model is able to get a probability for each patch of input images. For example, if the size of input image is $256\times256\times3$, and the size of patch for discriminator is $8\times8$, then the output of discriminator will be $32\times32\times1$, representing the score for each patch.
          
       For weight update of the GAN model, we used the similar way as WGAN, that is minimizing the Wasserstein distance between real distance and generated distance. In this model, a function $f_w$ with parameters $w$ are defined to construct the loss: 
       
       \begin{equation}
           \mathcal{L}_D = E_{x~P_r}[f_w(x)]- E_{x~P_g}[f_w(x)]
       \end{equation}
       
       where the $P_r$ is the distribution of real samples and $P_g$ is the distribution of generated samples, and $f_w$ is K-Lipschitz function.
       And the generator loss $L_G$ and discriminitor loss $L_D$ are:
       \begin{align}
           \mathcal{L}_G &= -E_{x~P_g}[f_w(x)] \\
           \mathcal{L}_D &= E_{x~P_g}[f_w(x)]-E_{x~P_r}[f_w(x)]
       \end{align}
            
       
       By minimizing $L_G$ and $L_D$ iteratively, we then can get a model that could generate visually plausible images.
       
    \paragraph{Step 2}
        Then in the next step, the GAN is further trained to generate images with target object conditioned on input position. A sequence of coordinate $[x1, y1, x2, y2]$ is first converted to a mask, where the corresponding pixels are set to one while the rest are zero. Then we fed it into an encoder and concatenate its feature with noise as the input of generator. 

        For the generated images, if they get relatively high score from discriminator, which means they are real enough and could be used for object detection, thus we can use a trained detection model to get the position of object from the generated images, compare it with the input bounding box data, and calculate an object position loss to generator as the generation constrain.  

        To achieve this, we applied a loss function similar to YOLO. It can be divided into three parts: bounding box regression loss, confidence loss and classification loss.
        
        We applied CIoU loss \cite{zheng2020distance} as bounding box regression loss:
        \begin{equation*}
            \mathcal{L}_{CIoU} = 1- IoU + \frac{\rho^2(B, B^{gt})}{c^2} + \alpha v
        \end{equation*}
         where $IoU$ is the naive Intersection over Union(IoU) metrics, $B$ and $B^{gt}$ denote the central points of detected bounding central points of detected bounding box and target bounding box, $\rho(\cdot)$ is the Euclidean distance and $x$ is the diagonal length of the smallest enclosing box covering the two boxes. $v$ refers to width-height consistency ratio,
        \begin{equation*}
            v = \frac{4}{\pi^2}(arctan\frac{w^{gt}}{h^{gt}} - arctan\frac{w}{h})^2
        \end{equation*}
        and $\alpha$ is a trade-off parameter defined as:
        \begin{equation*}
            \alpha = \frac{v}{(1-IoU)+v}
        \end{equation*}
        
        The confidence loss is used to determine whether there are objects in the prediction box, which is defined as below:
        \begin{equation}
            \mathcal{L}_{conf} = \sum_{i=0}^{S^{2}} \sum_{j=0}^{B} 1_{i j}^{\mathrm{obj}}\left(C_{i}-\hat{C}_{i}\right)^{2}+
            \lambda_{\text {noobj }} \sum_{i=0}^{S^{2}} \sum_{j=0}^{B} 1_{i j}^{\text {noobj }}\left(C_{i}-\hat{C}_{i}\right)^{2}
        \end{equation}
        
        The classification loss is used to determine whether the object in the prediction frame is the predicted object type:
        \begin{equation}
            \mathcal{L}_{clf} = \sum_{i=0}^{S^{2}} 1_{i}^{\mathrm{obj}} \sum_{c \in \text { classes }}
            \left(p_{i}(c)-\hat{p}_{i}(c)\right)^{2}
        \end{equation}
        
        Thus, the overall loss function is a linear combination between these three and can be written as follow:
        \begin{equation}
            \mathcal{L} = \alpha\mathcal{L}_{CIoU} + \beta\mathcal{L}_{conf} + \theta\mathcal{L}_{clf} 
        \end{equation}

    \subsection{Application}
        After training, we can feed the object position into this model to get augmented image together with the bounding box information. Moreover, we can generate some difficult training samples where object is small or object lies in the corner.
        
        With these generated images, we can construct an augmented dataset consists of both original images and generated images from our generator and retrain the object detection model on this augmented dataset.

\section{Experiments}
    \subsection{Experiment Setting}
    \paragraph{Dataset}
     We use VOC2007 as our dataset. It has totally 9k images where each image is labeled with several bounding boxes. We resize those images to $64\times64\times3$ before feeding into the discriminator. Given the coordinate of bounding box, we also generate mask of size $64\times64\times1$ to match the size of input images and then feed into the generator together. 

    \paragraph{GAN Model}
    We adopt Unet generator and N-Layer discriminator from Pix2Pix. We also adopt gradient penalty as WGAN-GP because of its advantages in both performance and stability. Generator and discriminator are trained iteratively. The goal of training generator is to generate indistinguishable samples for discriminator while the goal of training discriminator is to make sure it recognize real samples correctly. 
    
    \paragraph{Detection Model}
    We adopt YOLOv4 as our detection model, which is both fast and accurate. As is observed from our experiments, the YOLOv4 model pretrained on COCO dataset is also suitable for detection on VOC2007 dataset since 20 classes of VOC2007 are actually a subset of the 80 classes of COCO. So during training, we made use of pretrained YOLOv4 model to help us calculate the object position loss.

\subsection{Results Analysis}

\begin{figure}[htbp]
        \centering
        \begin{minipage}[t]{0.48\textwidth}
            \centering
            \includegraphics[scale=2]{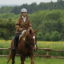}
        \end{minipage}
        \begin{minipage}[t]{0.48\textwidth}
        \centering
            \includegraphics[scale=2]{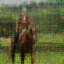}
        \end{minipage}
    \caption{Outcome of 1st-step model. \textbf{Left:} Real image from VOC dataset after resizing into $64\times64\times3$. \textbf{Right:} Corresponding generated image from our model.}
    \label{real_gen_compare}
    \end{figure}

    \paragraph{Step one}
    Results of the first step are shown in Figure \ref{real_gen_compare}, where the left image is real and the right one is generated. As can be seen, our model can generate similar image as real image. Although human being can recognize the object in generated image, i.e., a horse, a rider and the prairie background, there are still many noise-like pixels inside the generated image. These pixels may largely affect the performance of detection model. This phenomenon might due to the usage of WGAN, where the unsuitable magnitude of gradient penalty may distract its weights. Then the generator may not be able to concentrate fully on the competitive game itself, but also need to fulfill the requirement of gradient penalty item.
    
\begin{figure}
      \centering
      \includegraphics[scale=0.3]{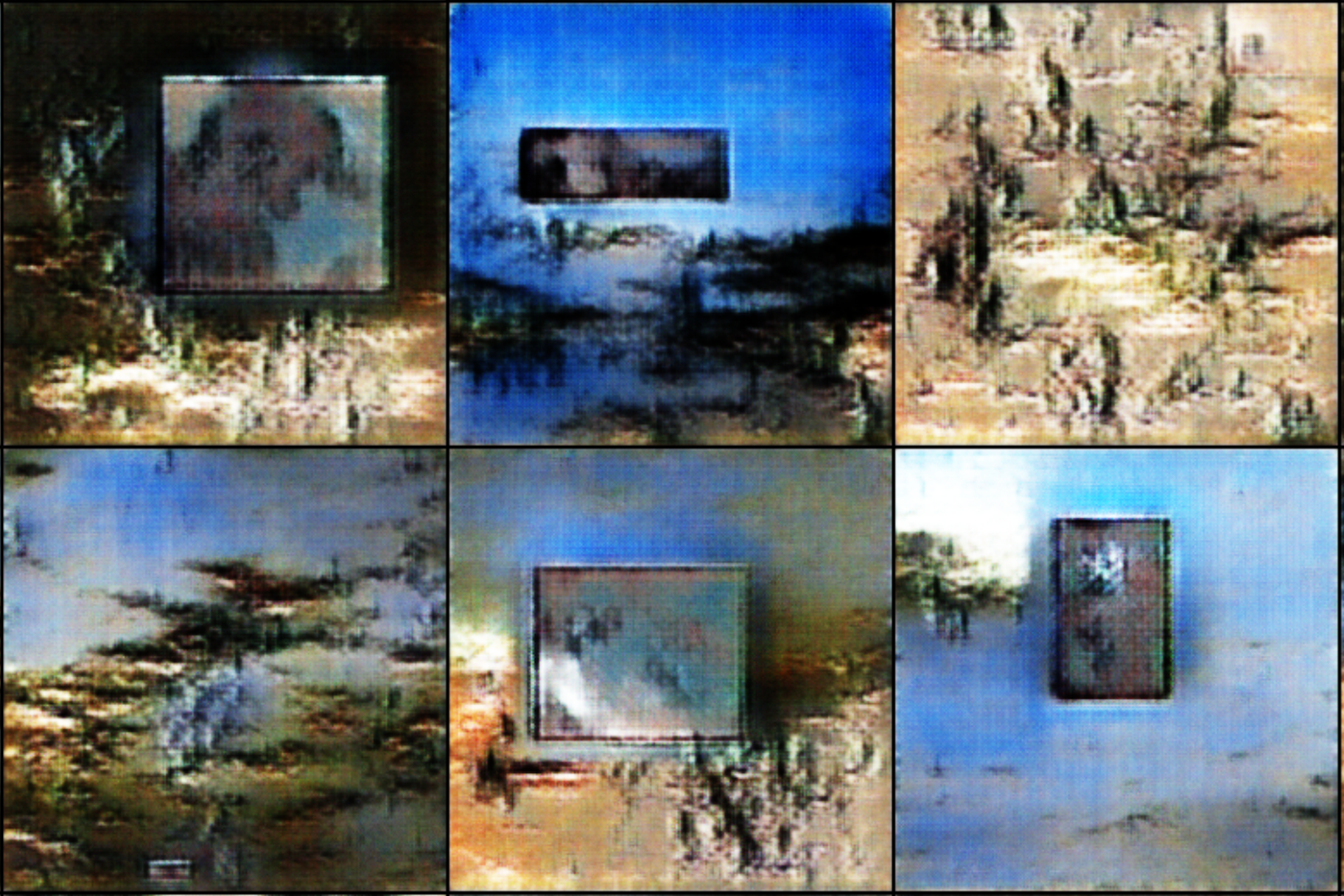}
      \caption{Temporary outcome of 2nd-step model.}
      \label{box_sample} 
     \end{figure}

    \paragraph{Step two} The results of second step are shown in Figure \ref{box_sample}. As shown in images, there is a rectangle in each generated image. This is because in the input of generator, we concatenate a mask to the noise in channel dimension. Without telling what generator shall do with this mask, it just automatically recognizes it as a shape it needs to display and thus there is an obvious boundary around the rectangle in each image. It is an interesting phenomenon, but it is actually not what we desired. In the future, one of our key directions is finding a better way to help generator understand the purpose of these input coordinates.
    
\subsection{Discussion}

    In this part, we will analyze our problems met during experiment. 
    
    It's well-known that GAN training is unstable. To address this, we've tried several methods to help stabilize the GAN training besides gradient penalty after our failure at the very beginning: learning rate decay after certain epochs and spectral normalization. Besides this, the largest problem is definitely about finding the balance between generator and discriminator.
    
\begin{figure}[htbp]
        \centering
        \begin{minipage}[t]{0.48\textwidth}
            \centering
            \includegraphics[scale=0.5]{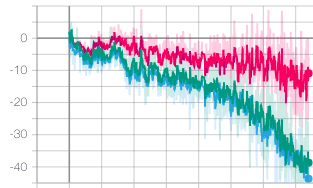}
        \end{minipage}
        \begin{minipage}[t]{0.48\textwidth}
        \centering
            \includegraphics[scale=0.6]{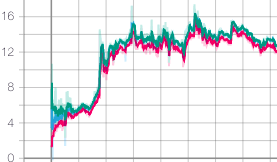}
        \end{minipage}
    \caption{Loss curve during our training. \textbf{Red:} score for real samples from discriminator. \textbf{Green and Blue:} score for fake samples from discriminator.}
    \label{training_curve}
    \end{figure}

    During training, we found that the discriminator outperforms generator most of the time even if we use the architectures proved to have similar capacity between generator and discriminator. The loss of such situation is shown in Figure \ref{training_curve}-left, where the generator cannot fool the discriminator after certain epochs. To address this, we use multiple tricks from internet, including adding Gaussian noise onto the input of discriminator, training generator with more steps at each iteration or just directly increasing the number of parameters of generator. However, it seems that this might relieve the problem to some degree, but the balance is still too fragile to maintain in our application. Most of the time, they compete with each other well at first, but after maybe several hours, things suddenly go wrong and the model just falls into the above situation that needs to stop and restart training.
    
    On the contrary, if we train the generator more frequently than discriminator, for example with a ratio of 3:1, then the loss curve will look like Figure \ref{training_curve}-right. It seems like this figure shows a perfect competition process, but actually, the generated images are even worse than the previous situation. The reason is that in this situation, the ability of discriminator is too poor to identify real images. A possible solution to these two situations is increasing both the number of discriminator steps and generator steps per iteration.
    
    What's more, compared to other application that use GAN for data augmentation, it seems that their GAN training is not so difficult as ours. One possible reason is that object detection dataset has more complex and distinct background between different images, which might confuse the generator a lot.
    
\subsection{Future Work}

    From our perspectives, this topic is interesting and creative. So we are going to continue on this after this module if possible.
    
    Our future work can be further extended in the following aspects:
    \begin{itemize}
        \item Continue on finding a better way to tell generator  the purpose of giving it those coordinates.
        \item Train a detection model based on augmented dataset to see if there are performance improvement compared to the case when only using original dataset.
        \item Further extend this framework, so that it can handle situation with multiple input coordinates and generate multiple objects on the generated image at the same time.

    \end{itemize}

\section{Conclusion}
    In this paper, we propose a novel framework to address one of the largest shortages of object detection task: lack of labeled training data. With the augmented dataset, we can further improve the performance of detection model intuitively.


\bibliographystyle{plain}
\bibliography{main}

\begin{thebibliography}{10}

\bibitem{antoniou2018data}
Antreas Antoniou, Amos Storkey, and Harrison Edwards.
\newblock Data augmentation generative adversarial networks, 2018.

\bibitem{arjovsky2017wasserstein}
Martin Arjovsky, Soumith Chintala, and Léon Bottou.
\newblock Wasserstein gan, 2017.

\bibitem{Bai_2018_ECCV}
Yancheng Bai, Yongqiang Zhang, Mingli Ding, and Bernard Ghanem.
\newblock Sod-mtgan: Small object detection via multi-task generative
  adversarial network.
\newblock In {\em Proceedings of the European Conference on Computer Vision
  (ECCV)}, September 2018.

\bibitem{2020arXiv200410934B}
Alexey {Bochkovskiy}, Chien-Yao {Wang}, and Hong-Yuan~Mark {Liao}.
\newblock {YOLOv4: Optimal Speed and Accuracy of Object Detection}.
\newblock {\em arXiv e-prints}, page arXiv:2004.10934, April 2020.

\bibitem{2014arXiv1412.7062C}
Liang-Chieh {Chen}, George {Papandreou}, Iasonas {Kokkinos}, Kevin {Murphy},
  and Alan~L. {Yuille}.
\newblock {Semantic Image Segmentation with Deep Convolutional Nets and Fully
  Connected CRFs}.
\newblock {\em arXiv e-prints}, page arXiv:1412.7062, December 2014.

\bibitem{2015arXiv150408083G}
Ross {Girshick}.
\newblock {Fast R-CNN}.
\newblock {\em arXiv e-prints}, page arXiv:1504.08083, April 2015.

\bibitem{2013arXiv1311.2524G}
Ross {Girshick}, Jeff {Donahue}, Trevor {Darrell}, and Jitendra {Malik}.
\newblock {Rich feature hierarchies for accurate object detection and semantic
  segmentation}.
\newblock {\em arXiv e-prints}, page arXiv:1311.2524, November 2013.

\bibitem{7112511}
Ross Girshick, Jeff Donahue, Trevor Darrell, and Jitendra Malik.
\newblock Region-based convolutional networks for accurate object detection and
  segmentation.
\newblock {\em IEEE Transactions on Pattern Analysis and Machine Intelligence},
  38(1):142--158, 2016.

\bibitem{2017arXiv170306870H}
Kaiming {He}, Georgia {Gkioxari}, Piotr {Doll{\'a}r}, and Ross {Girshick}.
\newblock {Mask R-CNN}.
\newblock {\em arXiv e-prints}, page arXiv:1703.06870, March 2017.

\bibitem{2015arXiv151203385H}
Kaiming {He}, Xiangyu {Zhang}, Shaoqing {Ren}, and Jian {Sun}.
\newblock {Deep Residual Learning for Image Recognition}.
\newblock {\em arXiv e-prints}, page arXiv:1512.03385, December 2015.

\bibitem{huang2018auggan}
Sheng-Wei Huang, Che-Tsung Lin, Shu-Ping Chen, Yen-Yi Wu, Po-Hao Hsu, and
  Shang-Hong Lai.
\newblock Auggan: Cross domain adaptation with gan-based data augmentation.
\newblock In {\em Proceedings of the European Conference on Computer Vision
  (ECCV)}, pages 718--731, 2018.

\bibitem{2018arXiv181204948K}
Tero Karras, Samuli Laine, and Timo Aila.
\newblock A style-based generator architecture for generative adversarial
  networks, 2019.

\bibitem{2020arXiv200313193L}
Kai {Li}, Yulun {Zhang}, Kunpeng {Li}, and Yun {Fu}.
\newblock {Adversarial Feature Hallucination Networks for Few-Shot Learning}.
\newblock {\em arXiv e-prints}, page arXiv:2003.13193, March 2020.

\bibitem{2017arXiv170802002L}
Tsung-Yi {Lin}, Priya {Goyal}, Ross {Girshick}, Kaiming {He}, and Piotr
  {Doll{\'a}r}.
\newblock {Focal Loss for Dense Object Detection}.
\newblock {\em arXiv e-prints}, page arXiv:1708.02002, August 2017.

\bibitem{2019arXiv191103029L}
Wei-Hong {Lin}, Jia-Xing {Zhong}, Shan {Liu}, Thomas {Li}, and Ge~{Li}.
\newblock {RoIMix: Proposal-Fusion among Multiple Images for Underwater Object
  Detection}.
\newblock {\em arXiv e-prints}, page arXiv:1911.03029, November 2019.

\bibitem{2015arXiv151202325L}
Wei {Liu}, Dragomir {Anguelov}, Dumitru {Erhan}, Christian {Szegedy}, Scott
  {Reed}, Cheng-Yang {Fu}, and Alexander~C. {Berg}.
\newblock {SSD: Single Shot MultiBox Detector}.
\newblock {\em arXiv e-prints}, page arXiv:1512.02325, December 2015.

\bibitem{NIPS2015_14bfa6bb}
Shaoqing Ren, Kaiming He, Ross Girshick, and Jian Sun.
\newblock Faster r-cnn: Towards real-time object detection with region proposal
  networks.
\newblock In C.~Cortes, N.~Lawrence, D.~Lee, M.~Sugiyama, and R.~Garnett,
  editors, {\em Advances in Neural Information Processing Systems}, volume~28.
  Curran Associates, Inc., 2015.

\bibitem{ronneberger2015unet}
Olaf Ronneberger, Philipp Fischer, and Thomas Brox.
\newblock U-net: Convolutional networks for biomedical image segmentation,
  2015.

\bibitem{2019arXiv191109070T}
Mingxing {Tan}, Ruoming {Pang}, and Quoc~V. {Le}.
\newblock {EfficientDet: Scalable and Efficient Object Detection}.
\newblock {\em arXiv e-prints}, page arXiv:1911.09070, November 2019.

\bibitem{vandeSandeICCV2011}
K.~E.~A. van~de Sande, J.~R.~R. Uijlings, T.~Gevers, and A.~W.~M. Smeulders.
\newblock Segmentation as selective search for object recognition.
\newblock In {\em IEEE International Conference on Computer Vision}, 2011.

\bibitem{990517}
P.~Viola and M.~Jones.
\newblock Rapid object detection using a boosted cascade of simple features.
\newblock In {\em Proceedings of the 2001 IEEE Computer Society Conference on
  Computer Vision and Pattern Recognition. CVPR 2001}, volume~1, pages I--I,
  2001.

\bibitem{2019arXiv191111929W}
Chien-Yao {Wang}, Hong-Yuan~Mark {Liao}, I-Hau {Yeh}, Yueh-Hua {Wu}, Ping-Yang
  {Chen}, and Jun-Wei {Hsieh}.
\newblock {CSPNet: A New Backbone that can Enhance Learning Capability of CNN}.
\newblock {\em arXiv e-prints}, page arXiv:1911.11929, November 2019.

\bibitem{2017arXiv170403414W}
Xiaolong {Wang}, Abhinav {Shrivastava}, and Abhinav {Gupta}.
\newblock {A-Fast-RCNN: Hard Positive Generation via Adversary for Object
  Detection}.
\newblock {\em arXiv e-prints}, page arXiv:1704.03414, April 2017.

\bibitem{2017arXiv171009412Z}
Hongyi {Zhang}, Moustapha {Cisse}, Yann~N. {Dauphin}, and David {Lopez-Paz}.
\newblock {mixup: Beyond Empirical Risk Minimization}.
\newblock {\em arXiv e-prints}, page arXiv:1710.09412, October 2017.

\bibitem{zheng2020distance}
Zhaohui Zheng, Ping Wang, Wei Liu, Jinze Li, Rongguang Ye, and Dongwei Ren.
\newblock Distance-iou loss: Faster and better learning for bounding box
  regression.
\newblock In {\em Proceedings of the AAAI Conference on Artificial
  Intelligence}, volume~34, pages 12993--13000, 2020.

\end{thebibliography}

\end{document}